\documentclass[11pt]{article}
\usepackage[utf8]{inputenc}
\usepackage[T1]{fontenc}
\usepackage[numbers,sort&compress]{natbib}
\usepackage{authblk}
\usepackage{geometry}
\usepackage{graphicx}
\usepackage{booktabs}
\usepackage{caption}
\usepackage{float}
\usepackage{amsmath}
\usepackage{longtable}
\usepackage{pdflscape}
\usepackage{array}
\usepackage{setspace}
\usepackage{hyperref}
\usepackage{tabularx}
\usepackage{multirow}
\usepackage{chngcntr}

\renewcommand{\arraystretch}{0.95}
\counterwithin{table}{section}

\title{\bfseries Multi-RADS Synthetic Radiology Report Dataset and Head-to-Head Benchmarking of 41 Open-Weight and Proprietary Language Models}

\author[1]{Kartik Bose}
\author[1]{Abhinandan Kumar}
\author[1]{Raghuraman Soundararajan}
\author[1]{Priya Mudgil}
\author[1]{Samonee Ralmilay}
\author[1]{Niharika Dutta}
\author[1]{Manphool Singhal}
\author[1]{Arun Kumar}
\author[2]{Saugata Sen}
\author[2]{Anurima Patra}
\author[2]{Priya Ghosh}
\author[3]{Abanti Das}
\author[4]{Amit Gupta}
\author[5]{Ashish Verma}
\author[6]{Dipin Sudhakaran}
\author[7]{Ekta Dhamija}
\author[8]{Himangi Unde}
\author[5]{Ishan Kumar}
\author[7]{Krithika Rangarajan}
\author[9]{Prerna Garg}
\author[8]{Rachel Sequeira}
\author[10]{Sudhin Shylendran}
\author[11]{Taruna Yadav}
\author[4]{Tej Pal}
\author[1,*]{Pankaj Gupta}

\affil[1]{Department of Radiodiagnosis, Postgraduate Institute of Medical Education and Research, Chandigarh, India 160012}
\affil[2]{Department of Radiodiagnosis, Tata Medical Center, Kolkata, India 700156}
\affil[3]{Department of Radiodiagnosis, All India Institute of Medical Sciences, Kalyani, India 741245}
\affil[4]{Department of Radiodiagnosis, National Cancer Institute, Jhajjar, India 124105}
\affil[5]{Department of Radiodiagnosis, Banaras Hindu University, Varanasi, India 221005}
\affil[6]{Department of Radiodiagnosis, Aster Malabar Institute of Medical Sciences, Kerala, India 670621}
\affil[7]{Department of Radiodiagnosis, All India Institute of Medical Sciences, New Delhi, India 110029}
\affil[8]{Department of Radiodiagnosis, Tata Main Hospital, Mumbai, India 400012}
\affil[9]{Department of Radiodiagnosis, Rajiv Gandhi Cancer Institute and Research Centre, Delhi, India 110085}
\affil[10]{Department of Radiodiagnosis, Baby Memorial Hospital, Kerala, India 670621}
\affil[11]{Department of Radiodiagnosis, All India Institute of Medical Sciences, Jodhpur, India 342005}
\affil[*]{Corresponding author: Pankaj Gupta --- \texttt{pankajgupta959@gmail.com}}

\date{}

\begin{document}
\maketitle

\begin{abstract}
Background: Reporting and Data Systems (RADS) standardize radiology risk communication; however, automated RADS assignment from narrative reports remains challenging due to guideline complexity and output-format constraints.

Purpose: To create a radiologist-verified synthetic multi-RADS benchmark (RXL-RADSet) and compare open-weight small language models (SLMs) with a proprietary model for RADS assignment.

Materials and Methods: RXL-RADSet contains 1,600 synthetic reports across 10 RADS frameworks and multiple modalities. Reports were generated by LLMs with simulated radiologist styles and two-stage radiologist verification. We evaluated 41 quantized SLMs (0.135--32B parameters) across 12 families and GPT-5.2 under guided prompting. Endpoints: validity and accuracy. Secondary sensitivity analysis compared guided vs zero-shot prompting.

Results: Under guided prompting, GPT-5.2: 99.8\% validity, 81.1\% accuracy (1,600 predictions). Pooled SLMs (65,600 predictions): 96.8\% validity, 61.1\% accuracy. Best open models (20--32B) achieved $\sim$99\% validity and up to 78\% accuracy. Guided prompting improved validity and accuracy relative to zero-shot.

Conclusion: RXL-RADSet is a radiologist-verified multi-RADS benchmark. Open-weight SLMs in the 20--32B range approach proprietary performance under guided prompting, but gaps remain for high-complexity tasks.

Code and data: \url{https://github.com/RadioX-Labs/RADSet}
\end{abstract}

\begin{figure}[H]
  \centering
  \includegraphics[width=0.9\textwidth]{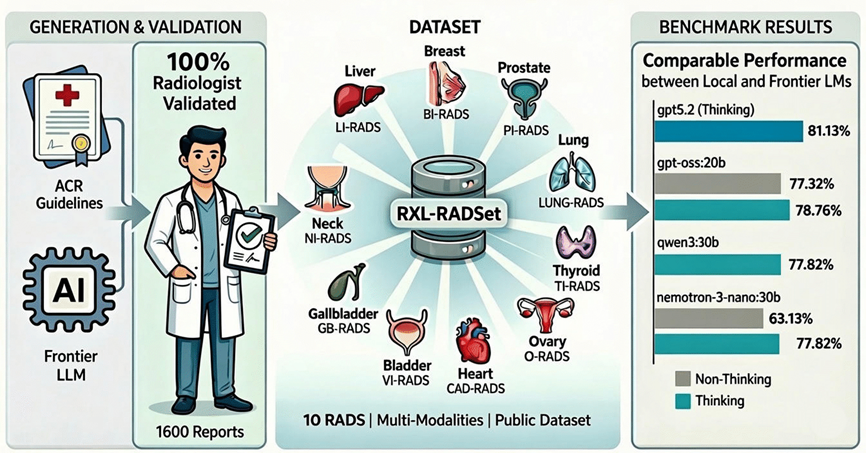}
  \caption{Overview of the RXL-RADSet Benchmarking Framework.}
\end{figure}

\section{Introduction}
Reporting and Data Systems (RADS) enable standardized risk stratification and communication across organ systems \citep{dorsi2013b,acr_rads,fraum2022}. Automating RADS category assignment from narrative radiology reports is increasingly valuable for integrating clinical decision support into reporting pipelines and secondary-use analytics \citep{wu2024,pandita2025synthetic}. Recent advances in natural language processing, particularly large language models (LLMs), have shown promise for medical report interpretation and structured extraction tasks across diverse radiology tasks \citep{kim2025benchmarking,bhayana2025,firoozeh2026,cozzi2024birads}.

Prior evaluations of LLMs in radiology have emphasized proprietary foundation models (for example, GPT-4/4o and Google's Gemini) and have reported strong performance in specific settings \citep{silbergleit2025,zheng2025}. At the same time there is growing interest in open-weight small language models (SLMs) as privacy-preserving, locally deployable alternatives \citep{pandita2025synthetic,magnini2025open,moll2025}. However, existing work has not comprehensively benchmarked model performance across the full spectrum of model scales, multiple RADS frameworks, and heterogeneous report styles.

Automated RADS scoring presents practical challenges: prompts must encode guideline logic and enforce strict output constraints because models can produce invalid or noncompliant outputs \citep{bhayana2025,mondal2025}; ground truth labeling requires expert adjudication \citep{savage2025open}; and credible evaluation requires broad clinical, modality, and stylistic diversity to avoid optimistic estimates from templated data \citep{nishio2024automatic}. To address these gaps, we introduce RXL-RADSet, a radiologist-verified synthetic benchmark spanning 10 RADS frameworks and 1,600 reports, and we benchmark 41 SLM configurations across 12 families (up to 32B parameters) alongside a proprietary reference model (GPT-5.2). We report both validity (schema-conformance) and accuracy (clinical correctness), stratified by model family, parameter scale, reasoning mode (Thinking vs Non-thinking), RADS scheme, and task complexity, and we analyze invalid-output modes and prompting sensitivity (guided vs zero-shot).

\section{Materials and Methods}

\subsection{Dataset: RXL-RADSet}
Dataset scope and structure. We constructed RXL-RADSet, a dataset of 1,600 synthetic radiology reports spanning 10 RADS frameworks across computed tomography (CT), magnetic resonance imaging (MRI), ultrasound (US), and mammography: breast (BI-RADS), coronary artery (CAD-RADS), gallbladder (GB-RADS), liver (LI-RADS), Lung-RADS, neck imaging (NI-RADS), ovary (O-RADS), prostate (PI-RADS), thyroid (TI-RADS), and urinary bladder (VI-RADS) \citep{cury2022,gupta2022} (Table~1, Figure~1, supplementary Table~1). Each report contained standard sections (Clinical Information/Indication, Technique, Findings, Impression). No real patient data were used.

\begin{table}[H]
\centering
\tiny
\caption{Per-domain RADS score statistics (counts).}
\begin{tabular}{lllllllllllllllllllllll}
\toprule
RAD & N & Modality & 0 & 1 & 2 & 2A & 2B & 3 & 4 & 4A & 4B & 4C & 5 & 6 & E & M & N & NP & NV & TIV & V \\
\midrule
BI-RADS & 100 & MRI & 0 & 9 & 23 & 0 & 0 & 16 & 19 & 0 & 0 & 0 & 13 & 20 & 0 & 0 & 0 & 0 & 0 & 0 & 0 \\
BI-RADS & 100 & US & 0 & 3 & 32 & 0 & 0 & 25 & 0 & 7 & 12 & 8 & 13 & 0 & 0 & 0 & 0 & 0 & 0 & 0 & 0 \\
BI-RADS & 100 & Mammo & 9 & 19 & 19 & 0 & 0 & 10 & 14 & 7 & 12 & 0 & 5 & 5 & 0 & 0 & 0 & 0 & 0 & 0 & 0 \\
GB-RADS & 100 & US & 8 & 29 & 33 & 0 & 0 & 14 & 6 & 0 & 0 & 0 & 10 & 0 & 0 & 0 & 0 & 0 & 0 & 0 & 0 \\
LI-RADS & 100 & CT & 0 & 17 & 5 & 0 & 0 & 37 & 3 & 0 & 0 & 0 & 34 & 0 & 0 & 2 & 0 & 0 & 0 & 2 & 0 \\
LI-RADS & 100 & CT/MRI & 0 & 0 & 0 & 0 & 0 & 0 & 0 & 0 & 0 & 0 & 0 & 1 & 0 & 1 & 14 & 63 & 0 & 21 & 0 \\
LI-RADS & 100 & MRI & 0 & 25 & 5 & 0 & 0 & 30 & 3 & 0 & 0 & 0 & 32 & 0 & 0 & 3 & 0 & 0 & 0 & 2 & 0 \\
LI-RADS & 100 & US & 0 & 31 & 30 & 0 & 0 & 39 & 0 & 0 & 0 & 0 & 0 & 0 & 0 & 0 & 0 & 0 & 0 & 0 & 0 \\
NI-RADS & 100 & CT & 0 & 55 & 1 & 4 & 1 & 38 & 1 & 0 & 0 & 0 & 0 & 0 & 0 & 0 & 0 & 0 & 0 & 0 & 0 \\
O-RADS & 100 & MRI & 1 & 1 & 39 & 0 & 0 & 22 & 12 & 0 & 0 & 0 & 25 & 0 & 0 & 0 & 0 & 0 & 0 & 0 & 0 \\
O-RADS & 100 & US & 2 & 3 & 49 & 0 & 0 & 25 & 5 & 0 & 0 & 0 & 16 & 0 & 0 & 0 & 0 & 0 & 0 & 0 & 0 \\
PI-RADS & 100 & MRI & 0 & 14 & 10 & 0 & 0 & 19 & 25 & 0 & 0 & 0 & 32 & 0 & 0 & 0 & 0 & 0 & 0 & 0 & 0 \\
TI-RADS & 100 & US & 0 & 3 & 9 & 0 & 0 & 17 & 23 & 0 & 0 & 0 & 33 & 0 & 0 & 0 & 15 & 0 & 0 & 0 & 0 \\
VI-RADS & 100 & MRI, CT & 0 & 9 & 28 & 0 & 0 & 14 & 21 & 0 & 0 & 0 & 21 & 0 & 0 & 0 & 7 & 0 & 0 & 0 & 0 \\
coronary-RADS & 100 & CT & 29 & 15 & 22 & 0 & 0 & 19 & 0 & 7 & 5 & 0 & 2 & 0 & 0 & 0 & 1 & 0 & 0 & 0 & 0 \\
LUNG-RADS & 100 & CT & 0 & 34 & 27 & 0 & 0 & 9 & 4 & 10 & 16 & 0 & 0 & 0 & 0 & 0 & 0 & 0 & 0 & 0 & 0 \\
\bottomrule
\end{tabular}
\end{table}

\subsection{Scenario planning and ground-truth labelling}
For each RADS framework we designed scenario templates reflecting modality, anatomy, lesion phenotype, risk strata, and decision points (see Supplementary Methods). Each scenario specified a target RADS category (including subcategories where applicable). The target served as ground truth after expert verification that report content supported that label.

\subsection{Synthetic report generation}
Reports were generated Dec 5--10, 2025 using proprietary LLMs (OpenAI GPT 5.2/5.1/4.1, Google Gemini 3 Pro/2.5 Pro, Anthropic Claude Sonnet/Haiku/Opus). We used provider chat UIs rather than APIs to preserve stylistic variability (no manual decoding-parameter tuning).

\subsection{Two-layer radiologist verification}
All reports underwent: (1) senior radiologist screening for realism and completeness; (2) subspecialty review for RADS adherence and label confirmation. Level-1 revisions ranged ~5--13.5\% by system; Level-2 revisions 3--8\% (see Supplementary Table~2).

\subsection{RADS complexity framework}
To quantify task difficulty we scored each RADS on three domains:
\begin{itemize}
  \item Categorization Burden (CB; 1--3): label granularity, modifiers.
  \item Algorithmic Workflow Depth (AWD; 1--3): decision-steps/branches.
  \item Interpretive Ambiguity Index (IAI; 1--4): subjectivity/acquisition dependence.
\end{itemize}
Total Complexity Score (TCS) was computed as
\[
\mathrm{TCS} = \mathrm{CB} + \mathrm{AWD} + \mathrm{IAI},
\]
range 3--10. Two radiologists scored each system; disagreements resolved by a third reviewer.

\subsection{Benchmark task: RADS extraction from report text}
Task: given the complete report text, models must output exactly one final RADS category. Prompting strategies:
\begin{itemize}
  \item Guided prompting (system + user): RADS-specific system prompt with rules, tiebreaks, and strict output constraints; user prompt: ``Read the report and generate final RADS category based on it.''
  \item Zero-shot prompting (user only): same user prompt without the system prompt. Zero-shot experiments were limited to three models (GPT-OSS 20B, Qwen3 30B, GPT-5.2) and five high-complexity RADS (LI-RADS CT/MRI, PI-RADS, O-RADS MRI/US).
\end{itemize}

\subsection{Thinking vs non-thinking variants}
For open-weight models that support it we evaluated:
\begin{itemize}
  \item Non-thinking: standard inference.
  \item Thinking-enabled: Ollama ``thinking'' flag enabled.
\end{itemize}
Only the final extracted category was scored; intermediate reasoning text was ignored after normalization.

\subsection{Models and inference setup}
SLM deployment. We evaluated 41 quantized SLMs (approx.\ 0.135--32B) across multiple families (Table~2), deployed locally (Ubuntu 22.04; Ollama; RTX 3090, 24\,GB). Quantization was required for local deployment; quantization formats and parameters are in Supplementary Table~2.

\begin{table}[H]
\centering
\scriptsize
\caption{Evaluated models, families, parameter scale, quantization, size, and prompting modes. M = millions; b = billions.}
\label{tab:models_full}
\begin{tabular}{p{3.6cm}p{1.1cm}p{2.6cm}p{1.9cm}p{1.4cm}p{1.4cm}}
\toprule
Model Family & Parameters & Quantization & Size & Prompt-guided: Thinking & Prompt-guided: Non-Thinking \\
\midrule
Qwen 3 & 0.6b & 4 bit (KM) & 522 MB & Yes & Yes \\
Qwen 3 & 1.7b & 4 bit (KM) & 1.4 GB & Yes & Yes \\
Qwen 3 & 4b & 4 bit (KM) & 2.5 GB & Yes & No \\
Qwen 3 & 8b & 4 bit (KM) & 5.2 GB & Yes & Yes \\
Qwen 3 & 14b & 4 bit (KM) & 9.3 GB & Yes & No \\
Qwen 3 & 30b & 4 bit (KM) & 18 GB & Yes & No \\
Qwen 3 & 32b & 4 bit (KM) & 20 GB & Yes & Yes \\
\midrule
Deepseek R1 & 1.5b & 4 bit (KM) & 1.1 GB & Yes & No \\
Deepseek R1 & 7b & 4 bit (KM) & 4.7 GB & Yes & No \\
Deepseek R1 & 8b & 4 bit (KM) & 5.2 GB & Yes & Yes \\
Deepseek R1 & 14b & 4 bit (KM) & 9 GB & Yes & Yes \\
Deepseek R1 & 32b & 4 bit (KM) & 19 GB & Yes & Yes \\
\midrule
GPT-OSS & 20b & 4 bit (MX) & 13 GB & Yes & Yes \\
\midrule
Ministral 3 & 3b & 4 bit (KM) & 3.0 GB & No & Yes \\
Ministral 3 & 8b & 4 bit (KM) & 6.0 GB & No & Yes \\
Ministral 3 & 14b & 4 bit (KM) & 9.1 GB & No & Yes \\
\midrule
Gemma 3 & 270m & 8 bit & 291 MB & No & Yes \\
Gemma 3 & 1b & 4 bit (KM) & 815 MB & No & Yes \\
Gemma 3 & 4b & 4 bit (KM) & 3.3 GB & No & Yes \\
Gemma 3 & 12b & 4 bit (KM) & 8.1 GB & No & Yes \\
Gemma 3 & 27b & 4 bit (KM) & 17 GB & No & Yes \\
\midrule
Nemotron 3 Nano & 30b & 4 bit (KM) & 24 GB & Yes & Yes \\
\midrule
Olmo 3 & 7b & 4 bit (KM) & 4.5 GB & Yes & No \\
Olmo 3 & 32b & 4 bit (KM) & 19 GB & Yes & No \\
\midrule
Llama4 & 16x17b & 4 bit (KM) & 67 GB & No & Yes \\
\midrule
Smollm2 & 360m & 16 bit (F) & 725 MB & No & Yes \\
Smollm2 & 1.7b & 8 bit & 1.8 GB & No & Yes \\
\midrule
Granite 4 & 350m & 16 bit (BF) & 708 MB & No & Yes \\
Granite 4 & 1b & 16 bit (BF) & 3.3 GB & No & Yes \\
Granite 4 & 3b & 4 bit (KM) & 2.1 GB & No & Yes \\
\midrule
Phi 4 & 14b & 4 bit (KM) & 11 GB & No & Yes \\
Phi 4 -- Reasoning & 14b-plus & 4 bit (KM) & 11 GB & No & Yes \\
\bottomrule
\end{tabular}
\end{table}

Proprietary reference model. GPT-5.2 was evaluated via the OpenAI Chat Completions full-precision API.

Inference-time notes. Measured inference times are not directly comparable across models due to single-GPU architecture and batch-loading multiple models into VRAM for throughput optimization.

\subsection{Deterministic decoding settings}
All extraction runs used deterministic settings: temperature=0.0, top\_p=1.0, top\_k=1, max completion tokens=50, seed=42, context window truncated to 10,000 tokens where applicable.

\subsection{Scoring and endpoints}
Validity: whether predicted label belongs to the RADS allowed set after normalization (whitespace/case/separator handling). Accuracy: effective (invalid counted incorrect; denominator=all cases) and conditional (valid-only). Main analyses used collapsed scoring (e.g., 4A/4B/4C collapsed to 4); exact subcategory matches also reported.

\subsection{Invalid-output taxonomy}
Invalid outputs were classified into five mutually exclusive categories: Missing, Out-of-format, Non-numeric / Multi-valued, Ambiguous, Other. For each model we counted invalids per category to explain gaps between effective and conditional accuracy.

\subsection{Statistical analysis}
All analyses were case-level. Primary stratifications: model family, parameter scale, reasoning mode, RADS stratum, and TCS. To estimate GPT-5.2 advantage we fit linear probability models with cluster-robust standard errors by case ID:
\begin{itemize}
  \item Overall adjusted model: outcome $\sim$ model group (GPT-5.2 vs open) + RADS stratum fixed effects.
  \item RADS-specific model: outcome $\sim$ model group + RADS stratum + model group$\times$RADS stratum.
\end{itemize}
P-values adjusted by Benjamini--Hochberg to control FDR; very small p reported as $p<0.001$. Analyses implemented in Python (pandas, matplotlib/seaborn) and statsmodels (cluster-robust SEs).

\section{Results}

\subsection{Dataset composition}
RXL-RADSet comprised 1,600 synthetic radiology reports spanning 10 RADS frameworks and four modalities (CT, MRI, mammography, ultrasound) (Supplementary Table 1). Counts were balanced for most schemes (n=100), with larger cohorts for BI-RADS (n=300), LI-RADS plus LR-TR (n=400), and O-RADS (n=200). RADS category prevalence was framework-specific (Table~1), motivating stratified reporting.

\subsection{RADS complexity: distribution and modality dependence}
Complexity scores (TCS) spanned 3--10 (mean 6.4) (Supplementary Table 3). The highest complexity was observed for LI-RADS CT/MRI (TCS 10) and PI-RADS (TCS 9), reflecting multi-step diagnostic matrices and higher interpretive ambiguity. Moderate-to-high complexity systems included O-RADS MRI/US, NI-RADS, and BI-RADS MRI (TCS 7--8). Lower complexity was observed for threshold-driven/screening systems such as Lung-RADS (TCS 4) and CAD-RADS (TCS 5). MRI-based systems were generally more complex than CT- or US-based systems.

\subsection{Main benchmark (guided prompting): performance across models and families}
Across all systems, validity ranged from 49.6\% (Gemma3 270M) to 100\% (Llama4 MoE), and effective (collapsed) accuracy ranged from 7.9\% (Gemma3 270M) to 81.1\% for GPT-5.2. Summary group-level results are shown in Table~\ref{tab:overall}. GPT-5.2 achieved near-perfect validity and the highest effective and conditional accuracies. Pooled open-source models achieved lower pooled validity and substantially lower effective accuracy.

\begin{table}[H]
\centering
\tiny
\caption{Overall performance under guided benchmarking: GPT-5.2 vs pooled open-source models.\label{tab:overall}}
\begin{tabular}{lrrrr}
\toprule
Group & N (predictions) & Valid (\%) & Effective acc., collapsed (\%) & Conditional acc., collapsed (\%) \\
\midrule
GPT-5.2 & 1,600 & 99.8 & 81.1 & 81.3 \\
Open-source models (pooled) & 65,600 & 96.8 & 61.1 & 63.1 \\
\bottomrule
\end{tabular}
\end{table}

\begin{figure}[H]
  \centering
  \includegraphics[width=0.92\textwidth]{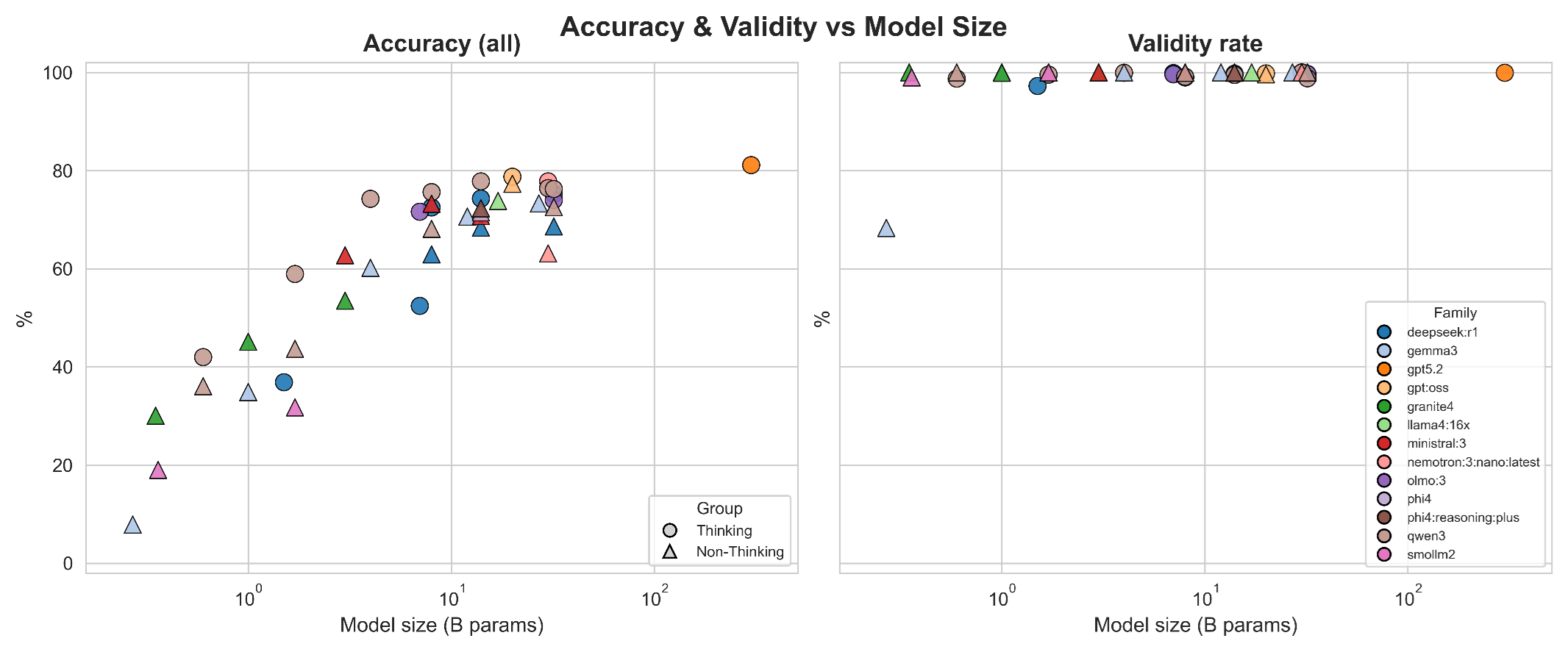}
  \caption{Accuracy and validity as a function of model size (fig2). Each point is a model; color = family; marker = reasoning mode.}
  \label{fig:size_valid_acc}
\end{figure}

Figure~\ref{fig:size_valid_acc} displays accuracy and validity versus model size (log scale), showing the concentration of top open-model performance in the 20--32B regime and GPT-5.2 as the far-right high-performing point.

\subsection{Reasoning mode effects (Thinking vs Non-thinking)}
Enabling thinking mode in supporting SLMs improved performance. Aggregated results by reasoning mode are presented in Table~\ref{tab:reasoning}. Thinking-mode open models showed higher validity and effective accuracy than their non-thinking counterparts; GPT-5.2 retained the highest absolute performance.

\begin{table}[H]
\centering
\tiny
\caption{Performance by reasoning mode (guided benchmark).\label{tab:reasoning}}
\begin{tabular}{lccc}
\toprule
Mode & Validity (\%) & Effective acc., collapsed (\%) & Conditional acc., collapsed (\%) \\
\midrule
Non-thinking (OS pooled) & 95.4 & 56.5 & 59.2 \\
Thinking (OS pooled) & 98.9 & 68.4 & 69.2 \\
GPT-5.2 (Thinking) & 99.8 & 81.1 & 81.3 \\
\bottomrule
\end{tabular}
\end{table}

\subsection{Invalid-output analysis: dominant failure modes}
Invalid outputs were most frequently classified as ``Other'', followed by Missing and Out-of-format (see Supplementary Table~6 for full breakdown). Smaller and instruction-misaligned models (e.g., Gemma3:270M, SmolLM2 variants) showed elevated missing and out-of-format rates, explaining much of their low effective accuracy despite occasionally reasonable conditional accuracy among their valid outputs.

\subsection{Scaling with model size and task difficulty}
Performance scaled strongly with model size. Table~\ref{tab:sizebin} shows binned results: validity rose from 82.9\% in the $\leq$1B bin to $\sim$99.2\% in the 10--29B bin, while effective collapsed accuracy rose from 27.0\% to $\sim$73.5\% across the same bins. GPT-5.2 exceeded the open-model plateau (81.1\% effective accuracy).

\begin{table}[H]
\centering
\tiny
\caption{Performance by model size bin (guided benchmark).\label{tab:sizebin}}
\begin{tabular}{lccc}
\toprule
Size bin & Validity (\%) & Effective acc., collapsed (\%) & Conditional acc., collapsed (\%) \\
\midrule
$\leq$1B & 82.9 & 27.0 & 32.6 \\
1--10B & 98.1 & 57.5 & 58.6 \\
10--29B & 99.2 & 73.5 & 74.1 \\
30--100B & 99.2 & 73.0 & 73.6 \\
GPT-5.2 & 99.8 & 81.1 & 81.3 \\
\bottomrule
\end{tabular}
\end{table}

\begin{figure}[H]
  \centering
  \includegraphics[width=0.92\textwidth]{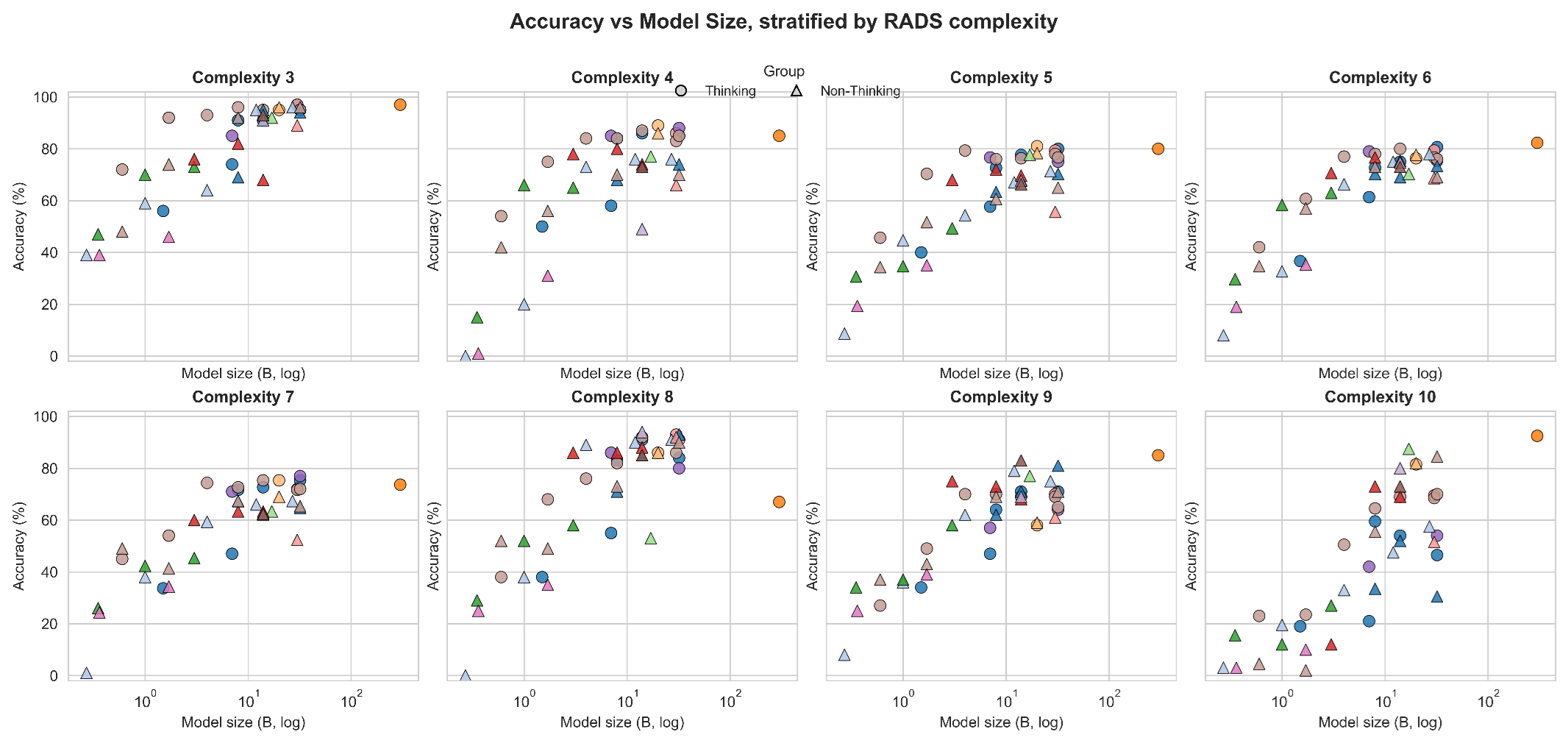}
  \caption{Accuracy versus model size stratified by RADS complexity (fig3).}
  \label{fig:complexity_accuracy}
\end{figure}

Figure~\ref{fig:complexity_accuracy} illustrates accuracy versus model size stratified by RADS complexity bins; accuracy declines and spread widens in higher-complexity strata, indicating greater sensitivity to model capacity and prompting.

\begin{figure}[H]
  \centering
  \includegraphics[width=0.92\textwidth]{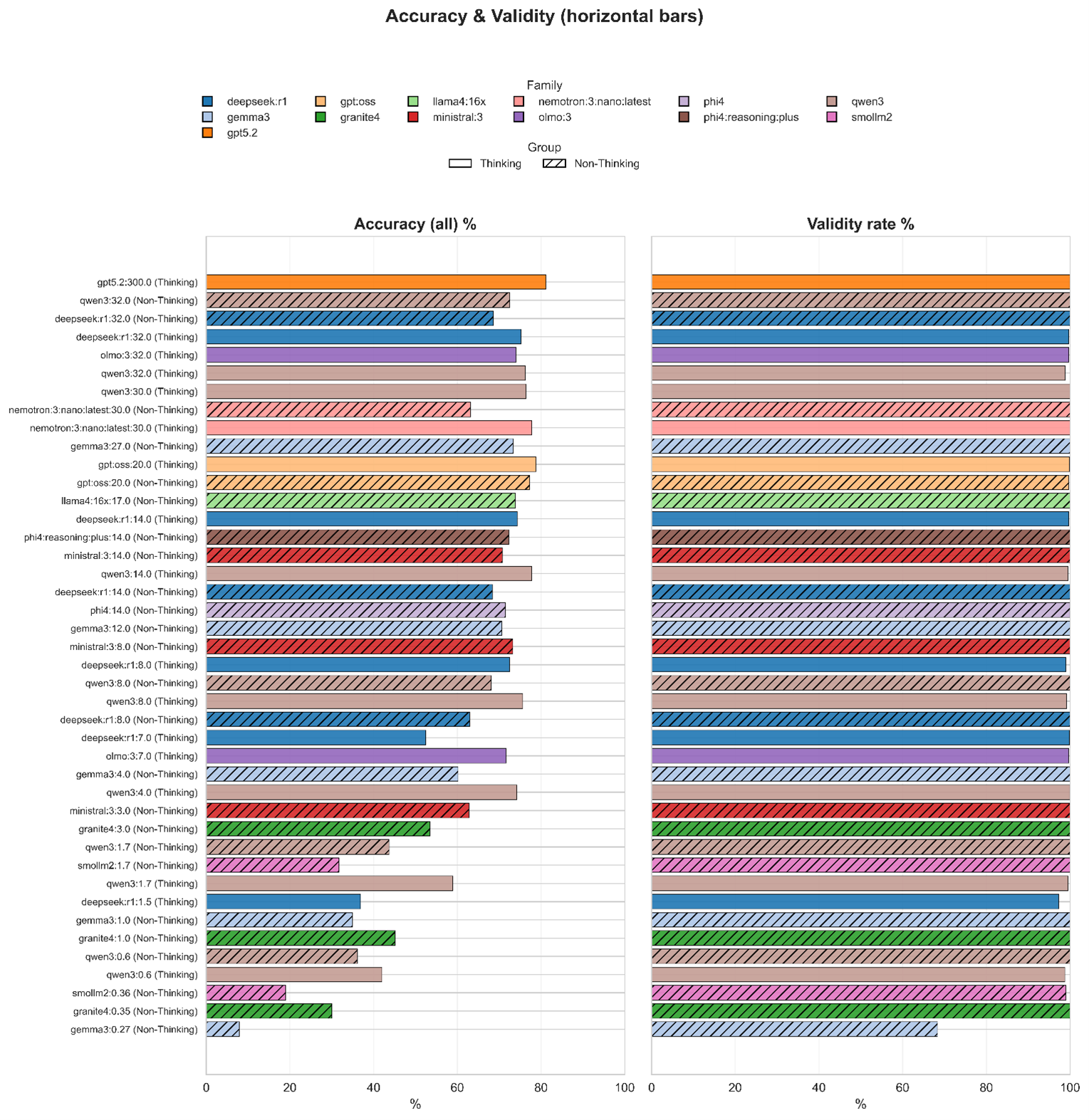}
  \caption{Model-wise accuracy and validity across evaluated systems (fig4).}
  \label{fig:by_system}
\end{figure}

Stratifying by RADS complexity (Supplementary Table~7; Figure~\ref{fig:by_system}) showed GPT-5.2 maintained high accuracy across complexity bins, whereas pooled SLMs' effective accuracy declined strongly in the highly complex bin (effective collapsed accuracy 49.4\%), despite high validity in that bin --- indicating classification difficulty rather than schema noncompliance as the dominant failure mode for complex tasks.

\subsection{Prompting sensitivity analysis: guided vs zero-shot}
In the restricted sensitivity analysis (three models; five high-complexity schemes), guided prompting improved pooled validity (99.2\% vs 96.7\%) and effective accuracy (78.5\% vs 69.6\%) relative to zero-shot (Table~\ref{tab:prompting}). The best-per-RADS analysis showed best-achievable validity 100\% in both modes, but best-achievable accuracy improved from 84.4\% (zero-shot) to 88.2\% (guided), driven largely by O-RADS MRI where an open model outperformed GPT-5.2 under guided prompting (Table~6).

\begin{table}[H]
\centering
\tiny
\caption{Pooled zero-shot vs guided prompting performance (subset).\label{tab:prompting}}
\begin{tabular}{lrrrrrr}
\toprule
Prompting mode & N & Valid (\%) & Invalid (\%) & Validity (\%) & Eff acc all (\%) & Cond acc valid (\%) \\
\midrule
Guided & 1,500 & 1,488 (99.2) & 12 (0.8) & 99.2 & 78.5 & 79.1 \\
Zero-shot & 1,500 & 1,450 (96.7) & 50 (3.3) & 96.7 & 69.6 & 72.0 \\
\bottomrule
\end{tabular}
\end{table}

\begin{figure}[H]
  \centering
  \includegraphics[width=0.92\textwidth]{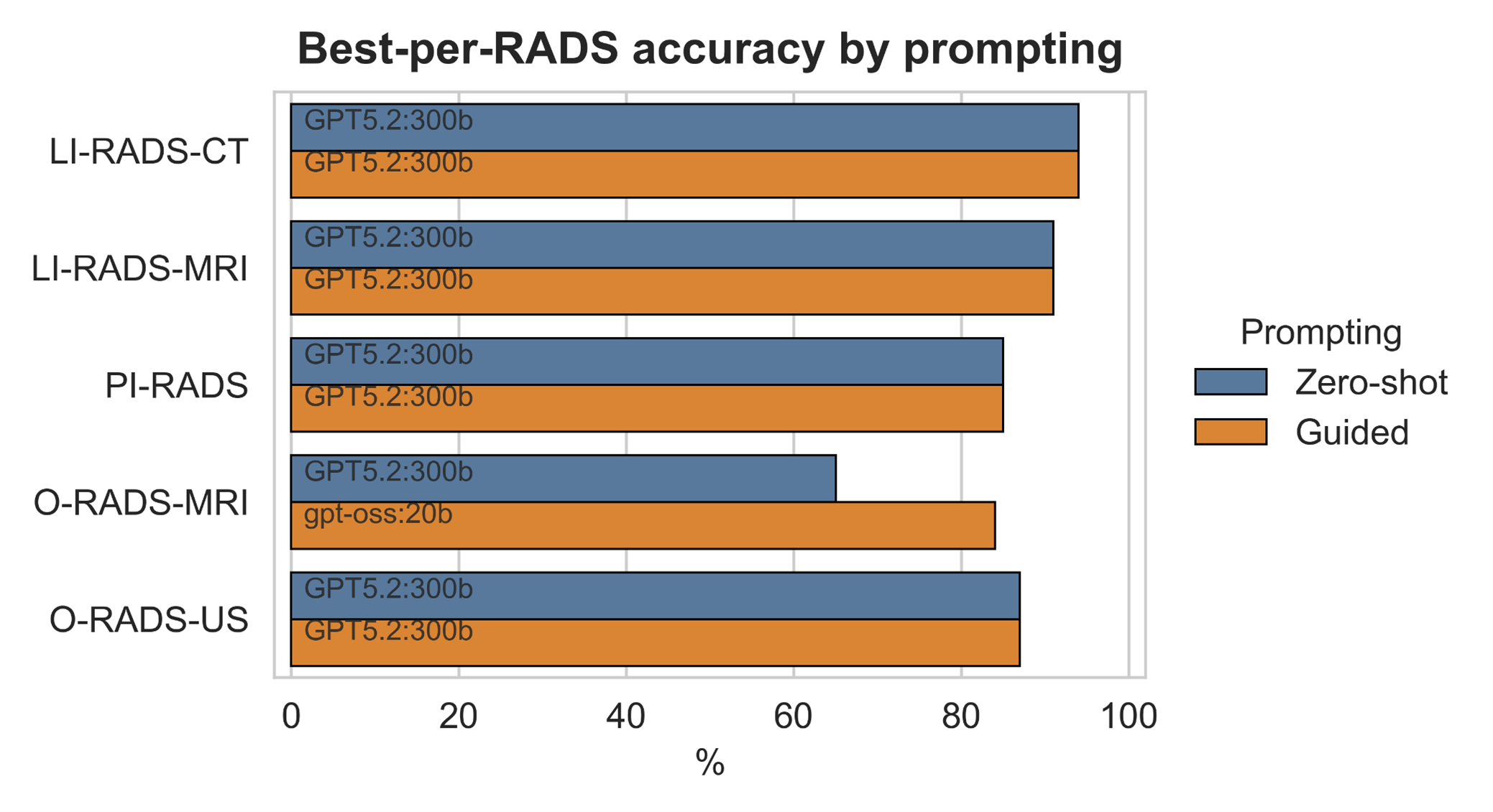}
  \caption{Best-per-RADS accuracy by prompting (fig5).}
  \label{fig:best_per_rads}
\end{figure}

\begin{figure}[H]
  \centering
  \includegraphics[width=0.92\textwidth]{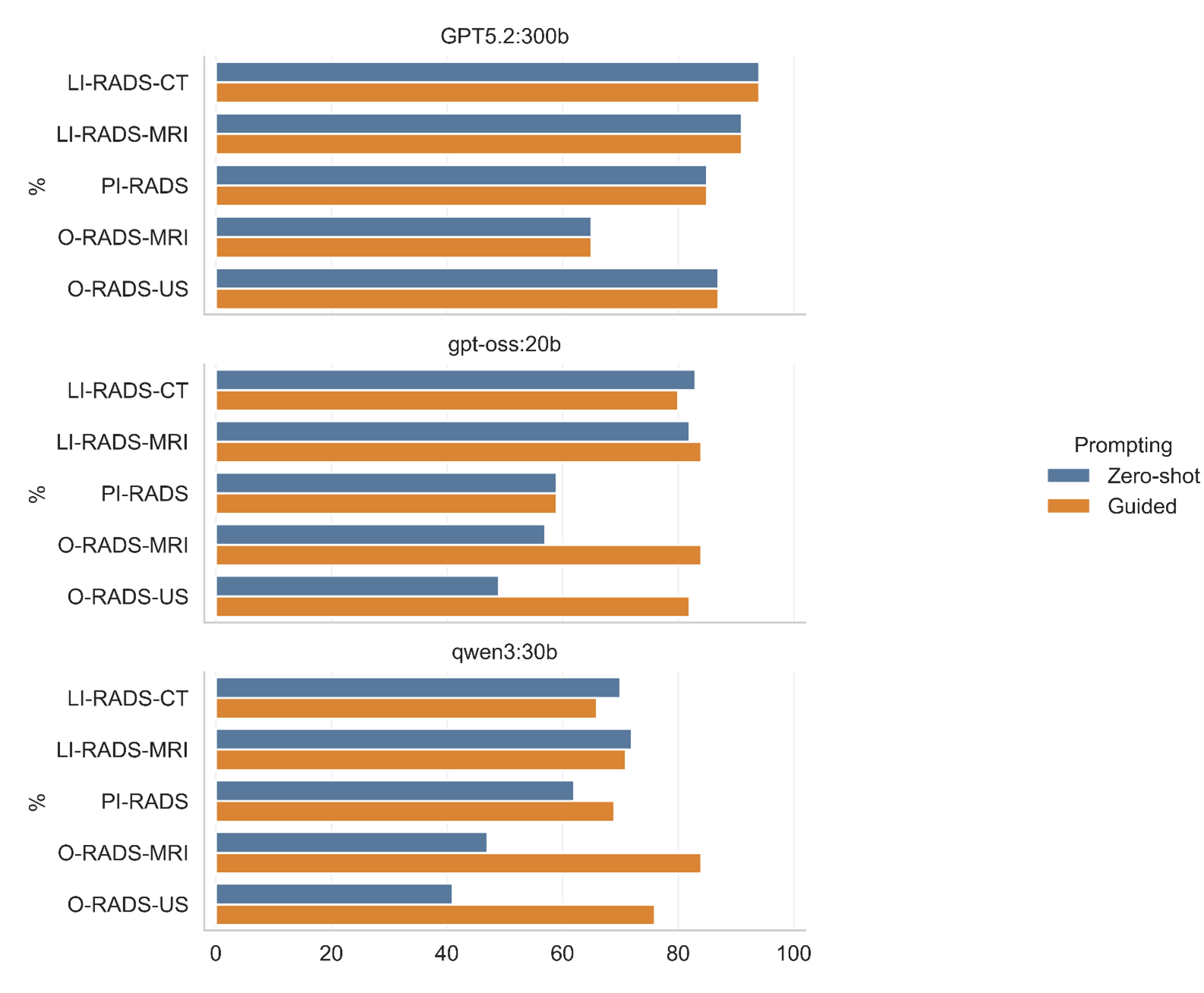}
  \caption{Per-RADS accuracy for selected models under zero-shot versus guided prompting (fig6).}
  \label{fig:per_rads_models}
\end{figure}

\subsection{Summary of key numeric contrasts}
Table~\ref{tab:overall} (above) summarizes the main group-level contrasts: GPT-5.2 exceeded pooled open-source models by +3.0 percentage points in validity and +20.0 percentage points in effective collapsed accuracy (absolute percentage-point differences). These contrasts remained significant in patient-level adjusted models with cluster-robust SEs (all contrasts $p<0.001$, FDR-adjusted).

\section{Discussion}

We introduced RXL-RADSet, a radiologist-verified synthetic benchmark of 1,600 reports spanning 10 RADS frameworks, and evaluated 41 open-weight SLM configurations (0.135--32B) alongside a proprietary reference (GPT-5.2). Three principal patterns emerged under guided prompting. First, there is a strong size--performance relationship with a marked inflection between sub-1B and $\geq$10B models, mirroring broader SLM scaling observations \citep{kim2025benchmarking,ranjit2024}. Second, mid-to-large open models attained near-ceiling validity but persistent classification gaps remained for high-complexity RADS, indicating that failures are often clinical reasoning errors rather than format noncompliance. Third, enabling reasoning scaffolds (Thinking) improved both validity and accuracy, showing that structured reasoning prompts provide benefits beyond enforcing output schema.

GPT-5.2 obtained the highest effective collapsed accuracy (81.1\%) with near-perfect validity (99.8\%), whereas the strongest open-weight models (20--32B) reached mid--high 70\% effective accuracy with $\sim$99\% validity. These results indicate that carefully prompted, large open models can approach frontier-model performance on structured RADS assignment while remaining amenable to local deployment and privacy-preserving workflows \citep{pandita2025synthetic,magnini2025open}. Still, the residual gap—most evident in the highest-complexity strata—suggests that improved clinical-rule application, not only increased model scale or formatting constraints, is required to close the remaining performance difference.

Our findings align with prior radiology LLM benchmarking showing proprietary models often lead overall while top open models follow closely \citep{kim2025benchmarking,cozzi2024birads,zheng2025}. They also illustrate that reported performance depends strongly on reporting style heterogeneity and prompt design: templated report settings yield higher agreement, whereas realistic heterogeneous narratives penalize smaller or less instruction-aligned models \citep{wu2024,moll2025}. This motivated our primary use of a guided prompt (system + user) to model realistic deployment practices; the sensitivity analysis confirmed that guidance materially improves both validity and accuracy versus zero-shot for the subset tested \citep{ferreira2024,wu2024}.

A practical implication is that prompt engineering alone may be necessary but not sufficient for the hardest RADS tasks. In some schemes guided prompts changed which model performed best (for example, O-RADS MRI shifted to a high-performing open model under guidance), indicating model-specific prompt sensitivity and supporting RADS-specific optimization or model routing strategies in deployment \citep{arnold2025performance,mondal2025}. For high-stakes clinical use, hybrid approaches—where LLM outputs are combined with deterministic rule application or used to extract features that feed a rule engine—can reduce harmful failures due to free-form or noncompliant outputs \citep{bhayana2025,mondal2025}.

Our invalid-output taxonomy showed that the dominant operational failure mode across many models is out-of-format or other schema noncompliance rather than ambiguous multi-valued responses; this explains substantial portions of the gap between conditional (valid-only) and effective accuracies in lower-performing families. Schema validation, post-processing, and calibrated abstention should therefore be prioritized as governance measures when integrating LLM-derived RADS outputs into clinical pipelines \citep{wang2025,das2025}. Such guardrails, together with human-in-the-loop review for high-risk categories, preserve auditability and patient safety \citep{prucker2025}.

Limitations. First, RXL-RADSet is synthetic despite rigorous radiologist verification; synthetic reports may not capture all real-world idiosyncrasies or documentation gaps. Second, open models were evaluated in quantized local deployments while GPT-5.2 was accessed at full precision, which may partially explain performance differences. Third, some RADS categories that are rare in practice remain underrepresented, which can limit precision for those classes. Fourth, inference-time comparisons are confounded by deployment choices (single GPU, VRAM batching) and are not reported as direct latency benchmarks.

Future work should evaluate (i) fine-tuning or instruction-tuning open models on radiology/RADS-specific corpora (potentially using synthetic + verified data) to determine how much of the residual gap is addressable by targeted adaptation \citep{ranjit2024,pandita2025synthetic}, (ii) hybrid pipelines that combine LLM-based feature extraction with deterministic RADS scoring, and (iii) prospective, multi-institutional validation on real-world reports to assess generalizability across reporting styles and languages \citep{cozzi2024birads,kim2025benchmarking}.

In summary, RXL-RADSet shows that SLMs---especially in the 20--32B range---can approach proprietary-model performance on structured RADS assignment under guided prompting, but frontier models retain an edge on the most complex decision-rule tasks. Operational deployment should emphasize deterministic output contracts, schema validation, and human oversight to ensure safety and reliability in clinical settings.

\section*{Data availability}
Code and dataset: \url{https://github.com/RadioX-Labs/RADSet}.

\clearpage
\appendix
\section*{Supplementary material}

\subsection*{General Plan for Simulated RADS Report Generation}
Global constraints for all synthetic reports
\begin{itemize}
  \item Each report includes:
    \begin{itemize}
      \item Key clinical information and relevant indication (age, screening/risk factors, clinical question).
      \item Technical description of examination, tailored to the target modality and system (e.g. MRI protocol, CT phases, US technique, etc.).
      \item Structured and realistic findings, including organ/system-specific morphology, region/zone descriptors, and lesion characterization per RADS logic.
      \item Lesion attributes: precise location, dimensions, imaging features relevant to malignancy or grading, and relation to adjacent structures.
      \item Impression: Clear, concise summary stating presence/absence and severity of abnormality, key scoring per RADS guidelines, and when warranted, a global/overall RADS category or risk assessment.
    \end{itemize}
  \item Each report was written according to:
    \begin{itemize}
      \item Content constraints and reporting logic of the relevant RADS.
      \item Representative risk scenarios, reflecting the clinical spectrum for that system.
    \end{itemize}
\end{itemize}

\subsection*{Simulated Radiologist Profiles}
\begin{itemize}
  \item Profile 1: Early-career generalist
  \item Profile 2: Mid-career generalist
  \item Profile 3: Recent/early-career subspecialist
  \item Profile 4: Senior subspecialist, high-volume expert
  \item Profile 5: Senior resident
\end{itemize}
For each profile, writing style, attention to detail, level of hedging, analytic depth, and use of standardized language were pre-defined.

\subsection*{Scenario Design}
Scenarios covered full risk spectrum, modality-appropriate patterns, complex cases (multifocality, adjacent-structure involvement), and system-specific complexities (NI-RADS dual scoring, LR-TR response categories, PI-RADS zone-specific logic).

\subsection*{LLM Prompting and Report Generation}
For each scenario/profile combination, prompts encoded style and core findings; reports generated by OpenAI, Anthropic, and Google models via chat UIs to preserve stylistic diversity.

\subsection*{Expert Review and Quality Assurance}
First-layer: senior radiologist review for realism. Second-layer: subspecialist verification and corrections. Initial QC used 10 test reports per RADS to refine generation protocols.

\subsection*{Prompting Strategies}
Prompt-guided (system + user): system prompts included precise category definitions, scoring algorithms, tie-break rules, and strict output format (single token category only). Zero-shot: minimal user prompt ("Read the report and generate final RADS category based on it").

\subsection*{Representative prompt excerpts}
BI-RADS (MRI) system prompt excerpt: (see manuscript repository for full prompt text) — instructs: expert breast radiologist, list categories, analyze morphology and kinetics, output ONLY category.

LI-RADS (US) system prompt excerpt: (see repository) — instructs: expert hepatologist/radiologist, list LI-RADS US categories, tiebreak rules, output ONLY category.

\subsection*{RADS complexity: method and scoring}
We evaluated 16 RADS systems and developed a rubric with three domains: CB (1–3), AWD (1–3), IAI (1–4). TCS = CB + AWD + IAI, range 3–10. Systems were reviewed against current lexicons (e.g., LI-RADS v2018, PI-RADS v2.1).


\clearpage
\section*{Supplementary Tables}

\begin{table}[H]
\centering
\scriptsize
\caption{Reports by system and modality (Supplementary Table 1).}
\begin{tabular}{lrrrrr}
\toprule
Domain & CT & MRI & Mammography & Ultrasound & Total \\
\midrule
BI-RADS & 0 & 100 & 100 & 100 & 300 \\
CAD-RADS & 100 & 0 & 0 & 0 & 100 \\
GB-RADS & 0 & 0 & 0 & 100 & 100 \\
LI-RADS* & 150 & 150 & 0 & 100 & 400 \\
Lung-RADS & 100 & 0 & 0 & 0 & 100 \\
NI-RADS & 100 & 0 & 0 & 0 & 100 \\
O-RADS & 0 & 100 & 0 & 100 & 200 \\
PI-RADS & 0 & 100 & 0 & 0 & 100 \\
TI-RADS & 0 & 0 & 0 & 100 & 100 \\
VI-RADS & 0 & 100 & 0 & 0 & 100 \\
\bottomrule
\multicolumn{6}{l}{\scriptsize *Includes LR-TR reports (50 = CT and 50 = MRI)}
\end{tabular}
\end{table}

\bigskip

\clearpage
\begin{landscape}
\scriptsize
\renewcommand{\arraystretch}{1.23}
\begin{longtable}{p{3.9cm} p{1.6cm} p{1.2cm} p{1.5cm} p{5.7cm} p{6cm}}
\caption{Quality assurance and revision of synthetic RADS reports (Supplementary Table 2).} \label{supp:table2_landscaped} \\
\toprule
\textbf{RADS system} & \textbf{Total reports} & \textbf{Revised by R1} & \textbf{Revised by R2} & \textbf{Typical Level 1 revision types} & \textbf{Typical Level 2 revision types} \\
\midrule
\endfirsthead

\multicolumn{6}{c}{{\bfseries Supplementary Table 2 (continued)}} \\
\toprule
\textbf{RADS system} & \textbf{Total reports} & \textbf{Revised by L1} & \textbf{Revised by L2} & \textbf{Typical Level 1 revision types} & \textbf{Typical Level 2 revision types} \\
\midrule
\endhead

\bottomrule
\endfoot

BI-RADS (Mammo)      & 100 & 12 & 6 & Clarified findings, improved impression & Added/matched category to correct lexicon \\
BI-RADS (MRI)        & 100 & 14 & 7 & Added lesion descriptors & Harmonized enhancement description \\
BI-RADS (US)         & 100 & 10 & 5 & Fixed clinical indication & Aligned scoring with latest guidance \\
PI-RADS              & 100 & 11 & 8 & Improved anatomical zone mapping & Adjusted DWI/T2 grading \\
LUNG-RADS            & 100 & 8  & 4 & Added nodule dimension & Risk category correction \\[1.3ex]
LI-RADS (CT/MRI)     & 100 each & 27 & 13 & Added CT phase details; clarified phase-dependent features & Refined observation feature interpretation; corrected category assignment \\[1.3ex]
LI-RADS (US)         & 100 & 5  & 3 & Specified screening indication & Category correction \\
LR-TR (post-treatment) & 100 & 10 & 6 & Corrected lexicon to v2024 & Corrected viability assignments \\[1.3ex]
GB-RADS              & 100 & 6  & 3 & Corrected technical description & Corrected GB-RADS scoring \\[1.3ex]
O-RADS (MRI, US)     & 100 each & 15 & 8 & Improved cystic/solid feature descriptions & Reassigned category per updated criteria \\[1.3ex]
NI-RADS              & 100 & 10 & 7 & Clarified anatomical region & Adjusted category per new system guidance \\[1.3ex]
TI-RADS              & 100 & 8  & 4 & Added nodule composition details & Refined risk scoring per guideline \\
VI-RADS              & 100 & 7  & 3 & Specified vesical region descriptors & Corrected category per imaging protocol \\[1.3ex]
CAD-RADS             & 100 & 6  & 3 & Clarified stenosis degree descriptions & Harmonized with latest CAD-RADS lexicon \\

\end{longtable}
\end{landscape}

\bigskip

\clearpage
\begin{landscape}
\setlength{\tabcolsep}{8pt}
\begin{longtable}{p{3.8cm} p{4.2cm} p{4.2cm} p{4.2cm} p{3.0cm}}
\caption{RADS complexity scoring (Supplementary Table 3)}\label{supp:table3} \\
\toprule
\textbf{RADS System} & \textbf{Categories (Count \& Score 1--3)} & \textbf{Assessment Workflow (Steps \& Score 1--3)} & \textbf{Diagnostic Ambiguity (Sources \& Score 1--4)} & \textbf{Total Complexity (1--10) \& Justification} \\
\midrule
\endfirsthead
\multicolumn{5}{c}{{\bfseries Table 3 (continued)}} \\
\toprule
\textbf{RADS System} & \textbf{Categories (Count \& Score 1--3)} & \textbf{Assessment Workflow (Steps \& Score 1--3)} & \textbf{Diagnostic Ambiguity (Sources \& Score 1--4)} & \textbf{Total Complexity (1--10) \& Justification} \\
\midrule
\endhead
\bottomrule
\endfoot

LI-RADS (CT/MRI) &
8+ (Score: 3): Includes LR 1-5, M, TIV, and NC categories. &
Score: 3: Multi-step: evaluate major features, apply table, then adjust via ancillary features. &
Score: 4: High; nuanced definitions of "washout" and "capsule" vary between phases. &
10/10 — Requires deep algorithmic knowledge and tie-breaking rules. \\[4pt]

PI-RADS &
5 (Score: 3): Standard 1-5 scale across different zones. &
Score: 3: Zone-dependent; requires switching between DWI (Peripheral) and T2 (Transition) as dominant. &
Score: 3: Moderate-High; interpretation of the "Symmetry" in TZ and post-biopsy artifacts. &
9/10 — High cognitive load due to the need for anatomical mapping and multi-sequence correlation. \\[4pt]

O-RADS (MRI) &
6 (Score: 3): Score 0–5; includes complex adnexal morphology. &
Score: 3: Requires detailed analysis of solid components and kinetic enhancement curves. &
Score: 2: Moderate; subjectivity in defining "solid" vs. "debris" in complex cysts. &
8/10 — Complexity is driven by the integration of morphology with dynamic contrast-enhanced (DCE) curves. \\[4pt]

LI-RADS (Post-LRT) &
4 (Score: 2): LR-TR Nonviable, Non-prog, Equivocal, or Viable. &
Score: 2: Focuses on mass like enhancement treatment bed. &
Score: 3: High; post-ablation/TACE inflammatory response often mimics viable tumor. &
7/10 — Challenging due to the "noise" of the treated environment and lack of size-based criteria. \\[4pt]

O-RADS (US) &
6 (Score: 2): 0–5 scale based on benign/malignant features. &
Score: 2: Morphological assessment combined with subjective "Color Score" for vascularity. &
Score: 3: Moderate; high inter-observer variability in assigning vascularity and wall thickness. &
7/10 — Balanced between morphology and Doppler; "Color Scoring" adds a layer of subjectivity. \\[4pt]

NI-RADS (CT) &
4 (Score: 2): Covers primary site and neck nodes (1–4). &
Score: 2: Assessment of enhancement patterns and mass effect in a post-treatment neck. &
Score: 3: High; post-radiation tissue changes significantly obscure early recurrence. &
7/10 — Relies on distinguishing expected post-op enhancement from suspect nodules. \\[4pt]

BI-RADS (MRI) &
7 (Score: 2): Standard 0–6 classification. &
Score: 2: Morphological description plus kinetic curve assessment (Washout/Plateau). &
Score: 3: Background parenchymal enhancement (BPE) can mask or mimic findings. &
7/10 — More complex than US/Mammo due to the fourth dimension (time/enhancement curves). \\[4pt]

VI-RADS &
5 (Score: 2): 1–5 scale for muscle invasion. &
Score: 2: Integration of T2, DWI, and DCE to "see" through the bladder wall layers. &
Score: 2: Moderate; exophytic growth vs. true wall invasion can be difficult to distinguish. &
6/10 — A specialized tool that requires high-resolution imaging to evaluate the muscularis propria. \\[4pt]

BI-RADS (Mammo) &
7 (Score: 2): Standard 0–6 classification. &
Score: 2: Systematic review of density, mass, calcifications, and asymmetries. &
Score: 2: Moderate; morphology of microcalcifications is a known point of observer drift. &
6/10 — The most established RADS; complexity is mitigated by extensive training and standardized lexicons. \\[4pt]

GB-RADS &
6 (Score: 2): 0-5 scale for gallbladder wall/masses. &
Score: 2: Evaluation of mucosal integrity, wall thickness, and polypoid features. &
Score: 2: Differentiating chronic cholecystitis/adenomyomatosis from early T1 cancer. &
6/10 — Newer system focusing on stratifying gallbladder wall thickening in non-acute setting. \\[4pt]

TI-RADS &
6 (Score: 1): Simple TR 1–5 and "Incomplete." &
Score: 3: Tedious point-summation across 5 distinct ultrasound categories. &
Score: 1: Low; the points system is highly prescriptive, reducing interpretation variability. &
5/10 — Low interpretive complexity but high clerical step count (adding points). \\[4pt]

BI-RADS (US) &
7 (Score: 1): Standard 0–6 classification. &
Score: 2: Evaluation of shape, margin, and echo pattern (parallel vs. non-parallel). &
Score: 2: Subjectivity in "circumscribed" vs. "microlobulated" margins. &
5/10 — Relies heavily on pattern recognition rather than complex multi-sequence logic. \\[4pt]

CAD-RADS &
6+ (Score: 1): Based on 0–5 stenosis grades. &
Score: 2: Calculation of stenosis percentage plus modifiers (S, V, G, P). &
Score: 2: Moderate; calcified plaques cause "blooming," leading to stenosis overestimation. &
5/10 — Primarily a measurement-based system; modifiers add secondary layers of detail. \\[4pt]

LUNG-RADS &
5+ (Score: 1): Levels 0-4 (with 4X/4S modifiers). &
Score: 2: Quantitative measurement of mean diameter and growth over time. &
Score: 1: Low; logic is largely driven by strict mm-thresholds for solid/subsolid nodules. &
4/10 — Highly objective and reproducible; complexity only increases in the presence of interval change. \\[4pt]

LI-RADS (US) &
3 (Score: 1): Negative, Sub-threshold, or Positive. &
Score: 1: Screening-based observation; identify presence of nodules/thrombus. &
Score: 1: Low; straightforward screening criteria for a high-risk population. &
3/10 — Designed for speed and high sensitivity in a screening setting, not detailed diagnosis. \\

\end{longtable}
\end{landscape}

\bigskip

\clearpage
\begin{landscape}
\setlength{\tabcolsep}{2pt} 
\begin{center}
\scriptsize
\renewcommand{\arraystretch}{0.95}
\begin{longtable}{p{4.0cm} r r r r p{2.4cm} p{2.4cm} p{2.4cm} p{2.4cm}}
\caption{Performance by model family (guided benchmark) (Supplementary Table 4)}\label{supp:table4_fixed} \\
\toprule
\textbf{Model family} & \textbf{N} & \textbf{Valid, n (\%)} & \textbf{Invalid, n (\%)} & \textbf{Validity \%} & \multicolumn{1}{c}{\textbf{Eff. acc., collapsed \%}} & \multicolumn{1}{c}{\textbf{Cond. acc., collapsed \%}} & \multicolumn{1}{c}{\textbf{Eff. acc., exact \%}} & \multicolumn{1}{c}{\textbf{Cond. acc., exact \%}} \\
\textbf{} & \textbf{} & \textbf{} & \textbf{} & \textbf{(95\% CI)} & \multicolumn{1}{c}{\textbf{(95\% CI)}} & \multicolumn{1}{c}{\textbf{(95\% CI)}} & \multicolumn{1}{c}{\textbf{(95\% CI)}} & \multicolumn{1}{c}{\textbf{(95\% CI)}} \\
\midrule
\endfirsthead

\multicolumn{9}{c}%
{{\bfseries Supplementary Table 4 (continued)}} \\
\toprule
\textbf{Model family} & \textbf{N} & \textbf{Valid, n (\%)} & \textbf{Invalid, n (\%)} & \textbf{Validity \%} & \multicolumn{1}{c}{\textbf{Eff. acc., collapsed \%}} & \multicolumn{1}{c}{\textbf{Cond. acc., collapsed \%}} & \multicolumn{1}{c}{\textbf{Eff. acc., exact \%}} & \multicolumn{1}{c}{\textbf{Cond. acc., exact \%}} \\
\textbf{} & \textbf{} & \textbf{} & \textbf{} & \textbf{(95\% CI)} & \multicolumn{1}{c}{\textbf{(95\% CI)}} & \multicolumn{1}{c}{\textbf{(95\% CI)}} & \multicolumn{1}{c}{\textbf{(95\% CI)}} & \multicolumn{1}{c}{\textbf{(95\% CI)}} \\
\midrule
\endhead

\bottomrule
\endfoot

GPT-5.2 (proprietary) & 1,600 & 1,597 (99.8) & 3 (0.2) & 99.8 (99.6--100.0) & 81.1 (79.2--83.1) & 81.3 (79.4--83.3) & 78.1 (76.0--80.1) & 78.2 (76.1--80.2) \\
GPT-OSS & 3,200 & 3,175 (99.2) & 25 (0.8) & 99.2 (98.9--99.6) & 78.0 (76.9--80.0) & 78.6 (77.5--80.7) & 74.4 (72.9--76.4) & 75.0 (73.5--77.0) \\
Llama4 16$\times$ & 1,600 & 1,600 (100.0) & 0 (0.0) & 100.0 (100.0--100.0) & 73.8 (71.8--76.8) & 73.8 (71.8--76.8) & 70.9 (68.9--73.6) & 70.9 (68.9--73.6) \\
Olmo3 & 3,200 & 3,182 (99.4) & 18 (0.6) & 99.4 (99.1--99.7) & 72.8 (71.1--75.0) & 73.2 (71.4--75.4) & 70.0 (68.2--71.9) & 70.4 (68.8--73.0) \\
Phi 4 & 1,600 & 1,565 (97.8) & 35 (2.2) & 97.8 (97.0--98.5) & 71.5 (69.7--74.1) & 73.1 (71.4--75.7) & 68.4 (66.4--70.9) & 69.9 (68.3--72.5) \\
Phi 4 (Reasoning+) & 1,600 & 1,599 (99.9) & 1 (0.1) & 99.9 (99.8--100.0) & 72.3 (70.4--74.7) & 72.4 (70.5--74.8) & 69.6 (67.6--72.1) & 69.6 (67.7--72.1) \\
Nemotron 3 Nano & 3,200 & 3,167 (99.0) & 33 (1.0) & 99.0 (98.5--99.3) & 70.5 (69.0--72.5) & 71.2 (69.7--73.2) & 67.6 (66.0--69.5) & 68.3 (66.7--70.2) \\
Ministral 3 & 4,800 & 4,764 (99.3) & 36 (0.8) & 99.3 (98.9--99.6) & 68.9 (67.7--70.8) & 69.4 (68.1--71.3) & 65.6 (64.2--67.6) & 66.1 (64.7--68.0) \\
DeepSeek R1 & 12,800 & 12,627 (98.6) & 173 (1.4) & 98.6 (98.4--98.9) & 63.9 (62.5--65.6) & 64.7 (63.4--66.5) & 60.9 (59.7--62.6) & 61.7 (60.6--63.4) \\
Qwen3 & 17,600 & 17,379 (98.7) & 221 (1.3) & 98.7 (98.5--99.0) & 63.8 (62.5--65.5) & 64.6 (63.4--66.3) & 61.3 (59.9--63.0) & 62.1 (60.8--63.7) \\
Granite4 & 4,800 & 4,527 (94.3) & 273 (5.7) & 94.3 (93.6--95.0) & 42.9 (41.1--44.9) & 45.5 (43.6--47.5) & 40.5 (38.8--42.3) & 42.9 (41.1--44.8) \\
Gemma3 & 8,000 & 7,089 (88.6) & 911 (11.4) & 88.6 (88.0--89.4) & 49.4 (48.1--51.0) & 55.7 (54.3--57.6) & 47.3 (46.0--49.2) & 53.4 (52.1--55.4) \\
SmolLM2 & 3,200 & 2,820 (88.1) & 380 (11.9) & 88.1 (86.6--89.5) & 25.4 (23.7--26.9) & 28.8 (27.1--30.8) & 24.5 (22.8--26.1) & 27.8 (26.1--29.8) \\
\end{longtable}
\end{center}
Effective accuracy counts invalid predictions as incorrect (denominator = all predictions). Conditional accuracy is computed among valid predictions only. `Collapsed' denotes major-category scoring; `Exact' requires exact label match after normalization.
\end{landscape}

\bigskip

\begin{table}[H]
\centering
\scriptsize
\caption{RADS-specific interaction modelling (Delta = GPT-5.2 minus open-source pooled) (Supplementary Table 5).}
\begin{tabular}{p{3.0cm}p{2.0cm}p{2.0cm}p{1.6cm}p{2.0cm}p{2.0cm}p{1.6cm}}
\toprule
RADS Scheme & $\Delta$ Validity & 95\% CI & Adj p-val (q) & $\Delta$ Eff Acc (Collapsed) & 95\% CI & Adj p-val (q) \\
\midrule
BI-RADS (Mammo) & +0.07\% & [--0.01\%, +0.16\%] & 0.09 & +21.1\% & [+14.4\%, +27.8\%] & $<$0.001 \\
BI-RADS (MRI) & +7.88\% & [+7.32\%, +8.43\%] & $<$0.001 & +27.8\% & [+20.3\%, +35.4\%] & $<$0.001 \\
BI-RADS (US) & +5.56\% & [+5.04\%, +6.08\%] & $<$0.001 & +16.0\% & [+8.7\%, +23.4\%] & $<$0.001 \\
CAD-RADS & +1.39\% & [--0.68\%, +3.47\%] & 0.19 & +19.8\% & [+16.6\%, +23.0\%] & $<$0.001 \\
GB-RADS & +1.85\% & [+1.63\%, +2.08\%] & $<$0.001 & +16.8\% & [+13.9\%, +19.7\%] & $<$0.001 \\
LI-RADS-CT & +0.12\% & [+0.02\%, +0.23\%] & 0.03 & +49.8\% & [+44.0\%, +55.6\%] & $<$0.001 \\
LI-RADS-LRTR & +0.15\% & [+0.01\%, +0.28\%] & 0.04 & +10.1\% & [+5.4\%, +14.9\%] & $<$0.001 \\
LI-RADS-MRI & +0.22\% & [+0.07\%, +0.37\%] & 0.007 & +45.7\% & [+40.1\%, +51.3\%] & $<$0.001 \\
LI-RADS-US & +0.05\% & [--0.02\%, +0.12\%] & 0.17 & +16.5\% & [+14.0\%, +18.9\%] & $<$0.001 \\
LUNG-RADS & +11.20\% & [+8.77\%, +13.62\%] & $<$0.001 & +18.6\% & [+12.8\%, +24.5\%] & $<$0.001 \\
NI-RADS & +6.90\% & [+5.24\%, +8.56\%] & $<$0.001 & --1.22\% & [--12.0\%, +9.6\%] & 0.82 \\
O-RADS-MRI & +2.61\% & [+2.49\%, +2.73\%] & $<$0.001 & --4.17\% & [--13.0\%, +4.7\%] & 0.38 \\
O-RADS-US & +2.51\% & [+2.35\%, +2.67\%] & $<$0.001 & +20.6\% & [+16.2\%, +24.9\%] & $<$0.001 \\
PI-RADS & +5.59\% & [+3.51\%, +7.66\%] & $<$0.001 & +26.3\% & [+20.1\%, +32.5\%] & $<$0.001 \\
TI-RADS & +1.98\% & [+1.60\%, +2.35\%] & $<$0.001 & +19.3\% & [+14.9\%, +23.8\%] & $<$0.001 \\
VI-RADS & +0.29\% & [+0.12\%, +0.46\%] & 0.001 & +17.0\% & [+11.5\%, +22.1\%] & $<$0.001 \\
\bottomrule
\end{tabular}
\end{table}

\bigskip

\begin{table}[H]
\centering
\scriptsize
\caption{Invalid output breakdown by model (Supplementary Table 6). \newline
\scriptsize{(* models operate in both thinking and non-thinking mode).}}
\begin{tabular}{p{3.4cm}rrrrrr}
\toprule
Model & Predictions & Invalid & Missing & Out-of-format & Multi-valued & Other \\
\midrule
GPT-5.2 & 1,600 & 3 & 0 & 0 & 0 & 3 \\
GPT-OSS:20b* & 3,200 & 25 & 2 & 1 & 0 & 22 \\
Llama4:16x17b & 1,600 & 0 & 0 & 0 & 0 & 0 \\
Olmo-3:32b & 1,600 & 7 & 1 & 0 & 0 & 6 \\
Olmo-3:7b & 1,600 & 11 & 2 & 0 & 0 & 9 \\
Phi-4:14b & 1,600 & 35 & 3 & 3 & 0 & 29 \\
Phi4-Reasoning+:14b & 1,600 & 1 & 0 & 0 & 0 & 1 \\
Nemotron-3-Nano:30b* & 3,200 & 33 & 2 & 1 & 0 & 30 \\
Ministral-3:14b & 1,600 & 10 & 1 & 0 & 0 & 9 \\
Ministral-3:8b & 1,600 & 13 & 1 & 0 & 0 & 12 \\
Ministral-3:3b & 1,600 & 13 & 1 & 0 & 0 & 12 \\
Deepseek-R1:1.5b & 1,600 & 13 & 2 & 0 & 0 & 11 \\
Deepseek-R1:7b & 1,600 & 10 & 2 & 0 & 0 & 8 \\
Deepseek-R1:8b* & 3,200 & 20 & 3 & 1 & 0 & 16 \\
Deepseek-R1:14b* & 3,200 & 40 & 4 & 2 & 0 & 34 \\
Deepseek-R1:32b* & 3,200 & 30 & 3 & 1 & 0 & 26 \\
Qwen3:0.6b* & 3,200 & 32 & 5 & 1 & 0 & 26 \\
Qwen3:1.7b* & 3,200 & 21 & 2 & 1 & 0 & 18 \\
Qwen3:4b & 1,600 & 10 & 2 & 0 & 0 & 8 \\
Qwen3:8b* & 3,200 & 25 & 4 & 1 & 0 & 20 \\
Qwen3:14b & 1,600 & 13 & 2 & 0 & 0 & 11 \\
Qwen3:30b & 1,600 & 2 & 0 & 0 & 0 & 2 \\
Qwen3:32b* & 3,200 & 30 & 3 & 2 & 0 & 25 \\
Granite4:350m & 1,600 & 20 & 2 & 1 & 0 & 17 \\
Granite4:1b & 1,600 & 25 & 3 & 1 & 0 & 21 \\
Granite4:3b & 1,600 & 22 & 2 & 0 & 0 & 20 \\
Gemma3:270m & 1,600 & 326 & 59 & 16 & 0 & 251 \\
Gemma3:1b & 1,600 & 55 & 7 & 2 & 0 & 46 \\
Gemma3:4b & 1,600 & 10 & 2 & 0 & 0 & 8 \\
Gemma3:12b & 1,600 & 12 & 2 & 0 & 0 & 10 \\
Gemma3:27b & 1,600 & 15 & 3 & 0 & 0 & 12 \\
SmolLM2:360m & 1,600 & 85 & 13 & 5 & 0 & 67 \\
SmolLM2:1.7b & 1,600 & 23 & 3 & 1 & 0 & 19 \\
\bottomrule
\end{tabular}
\end{table}

\bigskip

\begin{table}[H]
\centering
\scriptsize
\caption{Performance by RADS complexity bin (guided benchmark)  (Supplementary Table 7).}
\label{supp:table7_fixed}
\resizebox{\textwidth}{!}{%
\begin{tabular}{lp{2.6cm}r r r r r}
\toprule
Complexity bin & Group & N (predictions) & Valid, n (\%) & Validity \% (95\% CI) & Effective acc., collapsed \% (95\% CI) & Conditional acc., collapsed \% (95\% CI) \\
\midrule
\multirow{2}{*}{Minimally complex (<5)} 
 & OS models (pooled) & 8,200 & 7,698 (93.9) & 93.9 (92.7--94.9) & 73.5 (70.2--76.5) & 78.3 (75.0--81.3) \\
 & GPT-5.2 (reference) & 200 & 199 (99.5) & 99.5 (98.4--100.0) & 91.0 (86.7--94.8) & 91.5 (87.2--95.3) \\
\midrule
\multirow{2}{*}{Moderately complex (5--8)} 
 & OS models (pooled) & 45,100 & 43,739 (97.0) & 97.0 (96.8--97.2) & 62.1 (60.4--63.8) & 64.0 (62.3--65.8) \\
 & GPT-5.2 (reference) & 1,100 & 1,098 (99.8) & 99.8 (99.5--100.0) & 76.9 (74.4--79.3) & 77.0 (74.6--79.5) \\
\midrule
\multirow{2}{*}{Highly complex (>8)} 
 & OS models (pooled) & 12,300 & 12,057 (98.0) & 98.0 (97.3--98.7) & 49.4 (47.2--51.7) & 50.4 (48.1--52.8) \\
 & GPT-5.2 (reference) & 300 & 300 (100.0) & 100.0 (100.0--100.0) & 90.0 (86.4--93.4) & 90.0 (86.4--93.4) \\
\bottomrule
\end{tabular}%
}
\medskip

\scriptsize
\noindent Notes: Effective accuracy treats invalid outputs as incorrect (denominator = all predictions). Conditional accuracy is computed among valid outputs only. ``Collapsed'' denotes major-category scoring; ``Exact'' (not shown) requires exact subcategory match.
\end{table}

\bigskip

\begin{table}[H]
\centering
\scriptsize
\caption{Pooled zero-shot vs guided prompting performance (Supplementary Table 8).}
\begin{tabular}{lrrrrr}
\toprule
Prompting mode & N (predictions) & Valid, n (\%) & Invalid, n (\%) & Validity \% & Accuracy (effective, all) \% \\
\midrule
Guided & 1,500 & 1,488 (99.2) & 12 (0.8) & 99.2 & 78.5 \\
Zero-shot & 1,500 & 1,450 (96.7) & 50 (3.3) & 96.7 & 69.6 \\
\bottomrule
\end{tabular}

\medskip
\scriptsize{Notes: ``Accuracy (effective, all)'' treats invalid outputs as incorrect (denominator = all predictions). ``Accuracy (conditional, valid-only)'' (not shown here) is computed among valid outputs only.}
\end{table}

\clearpage

\bibliographystyle{unsrtnat}
\bibliography{manuscript}

\end{document}